\documentclass[10pt,twocolumn,letterpaper]{article}

\usepackage{iccv}
\usepackage{times}
\usepackage{epsfig}
\usepackage{graphicx}
\usepackage{amsmath}
\usepackage{amssymb}
\usepackage{multirow}
\usepackage{authblk}
\usepackage[breaklinks=true,bookmarks=false]{hyperref}

\iccvfinalcopy % *** Uncomment this line for the final submission

\def\iccvPaperID{****} % *** Enter the ICCV Paper ID here

\ificcvfinal\fi

\begin{document}

%%%%%%%%% TITLE
\title{Approximated Bilinear Modules for Temporal Modeling\vspace{-1.5ex}}

\author[1]{Xinqi Zhu}
\author[1]{Chang Xu}
\author[2]{Langwen Hui}
\author[2]{Cewu Lu}
\author[1]{Dacheng Tao}
\affil[1]{UBTECH Sydney AI Centre, School of Computer Science, 
Faculty of Engineering, \protect\\
The University of Sydney, 
Darlington, NSW 2008, 
Australia}
\affil[2]{Shanghai Jiao Tong University, Shanghai, China \protect\\
{\tt\small xzhu7491@uni.sydney.edu.au, 
\{c.xu, dacheng.tao\}@sydney.edu.au, \protect\\
\{kunosarges, lucewu\}@sjtu.edu.cn}\vspace{-1ex}
}

\maketitle
\ificcvfinal\thispagestyle{empty}\fi

%%%%%%%%% ABSTRACT
\begin{abstract}
    We consider two less-emphasized 
    temporal properties of video: 
    1. Temporal cues are fine-grained; 
    2. Temporal modeling needs reasoning. 
    To tackle both problems at once, we exploit 
    approximated bilinear modules (ABMs) for temporal modeling. 
    There are two main points making the modules effective: 
    two-layer MLPs can be seen as a constraint approximation 
    of bilinear operations, thus can be used to construct 
    deep ABMs in existing CNNs while reusing pretrained parameters; 
    frame features can be divided 
    into static and dynamic parts because of visual repetition 
    in adjacent frames, which enables temporal 
    modeling to be more efficient. 
    Multiple ABM variants and implementations are investigated, 
    from high performance to high efficiency. 
    %Based on them, we can keep the 
    %proposed modules both effective and efficient. 
    %The modules are also flexible enough to work with existing 
    %deep CNNs while reusing the pretrained parameters. 
    Specifically, we show how two-layer subnets in CNNs 
    can be converted to temporal bilinear modules 
    by adding an \emph{auxiliary-branch}. 
    Besides, we introduce snippet sampling and shifting inference 
    to boost sparse-frame video classification performance. 
    Extensive ablation studies are conducted to 
    show the effectiveness of proposed techniques. 
    Our models can outperform most state-of-the-art methods 
    on Something-Something v1 and v2 datasets without 
    Kinetics pretraining, and are also competitive 
    on other YouTube-like action recognition datasets. 
    Our code is available on https://github.com/zhuxinqimac/abm-pytorch.
\end{abstract}

%%%%%%%%% BODY TEXT
%------------------------------------------------------------------------
\section{Introduction}
Video action recognition has been one of the most fundamental problems 
in computer vision for decades. Since CNNs achieved great success in image
classification \cite{NIPS2012_4824,Simonyan2014VeryDC,
Szegedy2015GoingDW,He2016DeepRL,Huang2017DenselyCC}, 
deep models have been introduced 
to video domain for action recognition \cite{6165309,Simonyan2014TwoStreamCN,
Donahue2015LongTermRC,Tran:2015:LSF:2919332.2919929,
Wang2016TemporalSN,Sun2015HumanAR,
Feichtenhofer2016SpatiotemporalRN,NonLocal2018}. 
Different from image classification, 
video action recognition requires effort for temporal modeling, 
which is still an open problem in this field. 
%Due to the inherent high 
%complexity of videos, the progress of video understanding is much slower than 
%image classification.

Up to now there have been three promising ways 
for temporal modeling in action recognition. 
The first one is two-stream architecture \cite{Simonyan2014TwoStreamCN,
Feichtenhofer2016ConvolutionalTN,Wang2016TemporalSN} where the 
temporal information is captured by optical flow 
(can cost over $90\%$ of the run time \cite{Sun_2018_CVPR}). 
%which is 
%pre-calculated and usually computationally expensive to extract 
%(costs over $90\%$ of the whole run-time \cite{Sun_2018_CVPR}). 
The second one 
is 3D CNN \cite{6165309,Tran:2015:LSF:2919332.2919929,
Tran2017ConvNetAS,Carreira2017QuoVA}. 
%where the temporal cues 
%are distilled by deep neural networks. 
This methods has the problem of high pretraining cost because 
3D CNNs are hard to be directly used for small datasets due to overfitting. 
This pretraining can be very expensive, \eg 64 GPUs used in 
\cite{Carreira2017QuoVA} and 56 GPUs used in \cite{XieS3D}. 
%This method has the problem of high training cost 
%because to apply a 3D CNN to a (small) dataset, a very large video dataset 
%pretraining is usually needed otherwise we cannot get a mature model, 
%but this pretraining could be very expensive (64 GPUs are 
%used to train a model on Kinetics in \cite{Carreira2017QuoVA} and 56 GPUs are 
%used in \cite{XieS3D}). 
A late fusion step is usually used along with the above two methods for 
long-term prediction, which slowdowns their inference again. 
%Since the above two methods are usually trained 
%with short clips, a late fusion 
%step like TSN \cite{Wang2016TemporalSN} is needed for long-term prediction, 
%decreasing the inference speed again. 
We refer the above two methods as \emph{heavy} methods. 
On the contrary, the third way is 
a \emph{light} method which conducts temporal 
modeling based on 2D backbones 
where input frames are usually sparsely sampled 
\cite{Donahue2015LongTermRC,Ng2015BeyondSS,
Mahdisoltani2018FinegrainedVC,Shi2015ConvolutionalLN,
ECO_eccv18,Zhou2017TemporalRR,LinTSM}. 
%\eg recurrent models \cite{Donahue2015LongTermRC,Ng2015BeyondSS,
%Mahdisoltani2018FinegrainedVC,Shi2015ConvolutionalLN}, 
%late temporal convolutions \cite{ECO_eccv18}, 
%temporal relation network \cite{Zhou2017TemporalRR}, and 
%temporal shift model \cite{LinTSM}. 
Without expensive late fusion, preprocessing 
and postprocessing computational overheads are eliminated. 
%However, most \emph{light} methods have been shown less powerful than 
%\emph{heavy} methods in some 
%literatures \cite{Carreira2017QuoVA,Wang2018SMART}.
%Therefore a problem rises 
%that can we fill this unwanted gap by finding a 
%model that is as effective as \emph{heavy} 
%methods, while also as efficient as \emph{light} methods?
We value these merits of \emph{light} architectures, and discover a 
very powerful module which works harmoniously and effectively with 
them. Addtionally, the module we propose is very flexible and can also 
work with \emph{heavy} methods to get an evident performance boost.

%In this paper, we propose to adopt bilinear technique for temporal modeling 
%on top of 2D CNNs, and this idea comes from our two hypotheses about 
%temporal information. 
This paper is based on our two discoveries about videos. 
The first one is: \emph{Temporal cues are fine-grained}. Here we refer 
\emph{fine-grained} since the temporal information (motion or state changes) 
could be easily dominated by spatial information (color blobs). 
%\eg 2D color blobs can be more easily 
%captured by models than motion clues. 
This property can explain the usage of optical flow 
\cite{Simonyan2014TwoStreamCN} or dense trajectory \cite{Wang2013ActionRW} 
which magnify the impact of temporal features by extracting them explicitly. 
%3D CNNs can successfully 
%extract such fine-grained temporal cues from end-to-end learning due to 
%its high complexity and large-scale 
%video training dataset. 
%However, if we want to keep the 
%model light-weight, we need a cheaper method also being good at 
%extracting fine-grained information. 
As bilinear models have been shown effective for fine-graind classification 
\cite{lin2015bilinear,Lin2017ImprovedBP}, it motivates us to bring 
bilinear operation to video temporal modeling. 
%In \cite{lin2015bilinear}, 
%bilinear model has been successfully used for 
%fine-grained image classification, and it 
%inspired us to adopt this approach for temporal modeling. 
The second discovery is: \emph{Temporal modeling needs reasoning}. 
Unlike image processing where low-level features like texture or color blobs 
are crucial for classification, the key features in time could be more 
high-level and reasoning-required, \eg basic physics, causality, and human's 
intention. 
%Since the capability of reasoning state changes or motion dynamics has been 
%shown important for deriving the correct prediction 
%for a video \cite{Zhou2017TemporalRR}, we would like to find a more powerful 
%method to do reasoning than MLPs used in \cite{Zhou2017TemporalRR}. 
As the state-of-the-art technique for VQA problem which requires textual and 
visual reasoning is bilinear model \cite{Fukui2016MultimodalCB,Kim2017,
Yu2017MultimodalFB}, it again inspires us to use bilinear model to do 
reasoning for temporal sequences. 
%it inspired us to see if bilinear models are 
%effective for temporal reasoning in action recognition. 

Based on the discoveries above, we introduce our 
Approximated Bilinear Modules (ABMs) for temporal modeling. 
There are two insights that make ABMs effective. 
The first is that two-layer MLPs can be seen as a constrained 
approximation of bilinear operations, which enables us to flexibly construct 
ABMs inside existing deep networks while reusing pretrained parameters. 
The second is that adjacent frames are 
likely to be repetitive, so we propose to represent a frame feature with 
static and dynamic parts to achieve a more efficient computation. 
We investigate the module's multiple temporal variants, and how they 
can work with CNNs smoothly. Particularly, we introduce how 
ABMs can be carefully initialized so that they can be implanted into 
deep architectures while keeping pretrained parameters valid. 
In this paper, our proposed modules are instantiated 
with two backbones (2D-ResNet-34 and I3D) to 
show 1. the pure power of ABMs for temporal modeling, 
and 2. the complementarity with 3D networks. 
Besides, we also present a flow-inspired snippet sampling to bring 
short-term dynamics to \emph{light} models, 
and also introduce a \emph{shifting inference} protocol 
to further boost performance. 
Our models can outperform most previous state-of-the-art methods 
on Something-Something v1 and v2 datasets 
without large video dataset pretraining such as Kinetics, 
while keeping a decent accuracy-speed tradeoff.

%Based on the discoveries above, we introduce our 
%Approximated Bilinear Modules (ABMs) for temporal modeling. 
%Two of the temporal variants are 
%shown highly effective for temporal modeling, one of which adopts 
%the repetition property in adjacent frames 
%to further reduce learnable parameters. 
%In order to make our modules general and flexible, we introduce two 
%ways to embed our ABMs to deep architectures, both of which 
%can reuse the pretrained parameters and both are very effective for 
%video classification.  We present how ABMs can be carefully initialized 
%to keep pretrained parameters valid. 
%We instantiate our ABMs with two backbones (2D-ResNet-34 and I3D) to 
%show 1. the pure power of ABMs, and 2. the complementarity with 3D networks. 
%Additionally, we present a flow-inspired snippet sampling to bring 
%short-term dynamics to \emph{light} models, 
%and also introduce a \emph{shifting inference} trick 
%to further boost performance, with the cost of limited speed drop. 
%Our models can outperform all previous state-of-the-art methods 
%on Something-v1 and v2 datasets 
%without large video dataset pretraining such as Kinetics. The models 
%keep a decent accuracy-speed tradeoff, and the proposed ABMs are 
%shown highly effective for temporal modeling.

%------------------------------------------------------------------------
\section{Related Work}
{\bf Deep Learning for Action Recognition.} 
Nowadays deep neural networks have been popular for video action recognition 
\cite{Karpathy2014LargeScaleVC,Simonyan2014TwoStreamCN,
6165309,Tran:2015:LSF:2919332.2919929,Wang2018SMART,Donahue2015LongTermRC,
Ng2015BeyondSS,Wang2016TemporalSN,Carreira2017QuoVA,Zhou2017TemporalRR}. 
Karpathy \etal investigated 
deep models with various temporal fusion strategies on Sports-1M dataset 
\cite{Karpathy2014LargeScaleVC}. 
Ji \etal proposed 3D CNNs for 
end-to-end action recognition \cite{6165309}, and this idea has been 
extended to more general feature representation 
learning by C3D \cite{Tran:2015:LSF:2919332.2919929}. 
Later, more powerful and deeper 3D CNNs with 
variations have been introduced, such as Res3D 
\cite{Tran2017ConvNetAS}, I3D 
\cite{Carreira2017QuoVA} (using inflated ImageNet-pretrained parameters), 
S3D \cite{XieS3D} (looking for cheaper 3D convolutions), 
and ARTNet \cite{Wang2018SMART}. 
Usually 3D architectures are heavy and reuqire expensive pretraining. 
%A drawback of 3D models is that 
%the architectures are usually heavy and of high complexity, requiring 
%large-scale video datasets for pretraining, which is computationally 
%expensive. 
Two-Stream architecture \cite{Simonyan2014TwoStreamCN} 
utilizes pre-extracted 
optical flow to capture temporal information. Feichtenhofer \etal 
investigated different fusion methods to more efficiently conduct 
two-stream processing \cite{Feichtenhofer2016ConvolutionalTN}. 
%But a problem 
%of two-stream model is that the extraction of optical flow is not cheap, 
%being a bottleneck for its efficiency. 
For long-term temporal modeling, Donahue \etal \cite{Donahue2015LongTermRC}, 
Ng \etal \cite{Ng2015BeyondSS} and Shi \etal \cite{Shi2015ConvolutionalLN} 
adopted LSTMs in video action recognition. Later, Ballas \etal 
\cite{Ballas_2016} proposed a ConvGRU for video understanding using 
multi-layer feature maps as inputs. Later Wang \etal \cite{Wang2016TemporalSN} 
proposed TSN architecture for long-term 
modeling, which is popular in 3D-based methods. 
Recently some works focus on light-weight temporal modeling. 
Zhou \etal \cite{Zhou2017TemporalRR} to 
do late reasoning in action recognition. 
Zolfaghari \etal \cite{ECO_eccv18} introduced ECO, a 
hybrid architecture of BN-Inception and 3D-ResNet-18 for fast 
action recognition. By shifting part of feature vectors, Lin \etal 
\cite{LinTSM} proposed TSM to do temporal modeling without extra parameters. 
%These long-term methods are light-weight and can usually run efficiently, 
%but they may have less temporal capacity than 
%3D models \cite{Carreira2017QuoVA,Wang2018SMART}. 
Our work is to implant the bilinear operations into normal convolutions, 
with the goal of exploiting the fine-grained nature of temporal dynamics, 
while reusing the normal pretrained parameters as well.
%In this paper, we focus on developing a model that keeps 
%efficiency as \emph{light} methods, while increasing temporal reasoning 
%capacity and producing comparative performance as \emph{heavy} methods.

{\bf Bilinear Models.} 
Bilinear pooling has been promising in many computer vision tasks 
\cite{lin2015bilinear,Gao2016CompactBP,
Fukui2016MultimodalCB,Kim2017,Yu2017MultimodalFB,
Lin2017ImprovedBP,Yu2018HierarchicalBP,Gatys2016ImageST}
Lin \etal utilized two 
streams of CNNs for fine-grained classification by extracting 
two branches of features 
and fusing them with outer product \cite{lin2015bilinear}. 
To address the high-dimension problem 
of bilinear pooling, Gao \etal introduced compact 
bilinear pooling \cite{Gao2016CompactBP} 
where the projected low-dimensional feature's kernel 
approximates the original polynomial kernel. 
In VQA tasks where inputs are naturally bi-modal, bilinear 
models are shown very effective 
\cite{Fukui2016MultimodalCB,Kim2017,Yu2017MultimodalFB}. 
Kim \etal \cite{Kim2017} brought bilinear 
low-rank approximation \cite{NIPS2009_3789} to VQA, and 
Yu \etal \cite{Yu2017MultimodalFB} proposed a rank-n variant. 
The bilinear operation has also been shown effective to 
pool over hierarchical layers in CNNs \cite{Yu2018HierarchicalBP}. 
%This technique has been successfully adopted in VAQ tasks 
%\cite{Fukui2016MultimodalCB}, showing 
%superior performance over other multimodal fusion methods. 
%Later, a more efficient and flexible approach, multimodal low-rank 
%bilinear pooling (MLB) \cite{Kim2017}, 
%has been proposed. 
%This model approximated the weight tensor in bilinear pooling with rank-1 
%3-order tensors, significantly reducing the computational complexity. 
%Later Yu \etal introduced a more general rank-n variant of 
%MLB \cite{Yu2017MultimodalFB} for 
%more robust performance and faster convergence.

There have been some attempts to apply bilinear approaches to video 
action recognition \cite{Diba2017DeepTL,Wang2017SpatiotemporalPN,
Girdhar2017AttentionalPF}. Diba \etal \cite{Diba2017DeepTL} 
proposed Temporal Linear Encoding (TLE) to encode a video into a compact 
representation using compact bilinear pooling \cite{Gao2016CompactBP}. 
%However their method was not designed for long-term temporal 
%modeling. 
Wang \etal \cite{Wang2017SpatiotemporalPN} 
introduced a spatiotemporal pyramid architecture 
using compact bilinear pooling to fuse temporal and 
spatial information with attention. In 
\cite{Girdhar2017AttentionalPF}, a top-down attention model has been 
developed based on the rank-1 approximation of bilinear pooling operation. 
However, these attempts either apply the bilinear operations to 
spatial features or fuse multiple branches of modalities, but none 
show its potential for temporal modeling. 
%Unlike the attempts mentioned above, we focus on developing 
%bilinear-based modules with high flexibility for temporal 
%modeling, in a plug-and-play manner while keeping decent efficiency.

%------------------------------------------------------------------------
\section{Approach}
\begin{figure*}[t]
\begin{center}
   \includegraphics[width=0.95\linewidth]{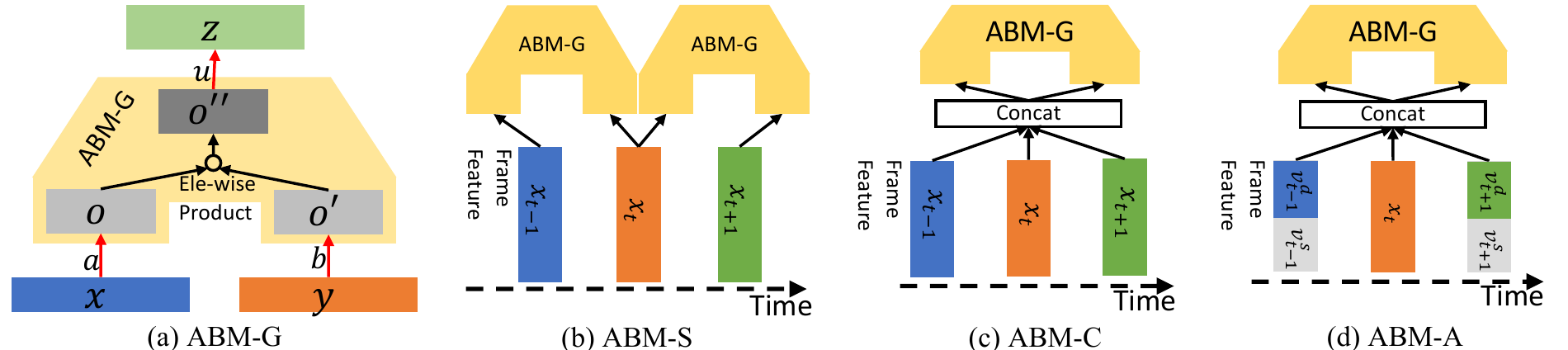}\vspace{-1.5ex}
\end{center}
    \caption{ABM variants. (a) ABM-G. (b) ABM-S. (c) ABM-C. (d) ABM-A, where 
    $\boldsymbol{x}_{t} = Concat(\boldsymbol{v}^{s}_{t}, 
    \boldsymbol{v}^{d}_{t})$.}
\label{fig:ABMs}\vspace{-1.5ex}
\end{figure*}

We first give definitions of ABM with variants, then introduce 
how they can work with deep architectures smoothly. Later we 
present two instantiations of our modules, snippet sampling, 
and implementation details.

\subsection{General Approximated Bilinear Module} \label{sec:abm_define}
{\bf Definition.} 
%A bilinear pooling module \cite{Tenenbaum2000SeparatingSA,lin2015bilinear} 
%calculates the Gram matrix of two global descriptors and maps the vectorized 
%matrix to an output vector containing pair-wised relation information 
%with massive learnable parameters. 
A bilinear pooling module \cite{Tenenbaum2000SeparatingSA,lin2015bilinear} 
calculates the Gram matrix of two global descriptors to learn 
a pair-wised relation feature. 
In this paper, we ignore the pool-over-location operation in bilinear 
pooling, but focus on the simpler bilinear module taking 
two vectors as inputs:
\begin{align}
\boldsymbol{z} = \boldsymbol{W} Vec( 
    \boldsymbol{x}\boldsymbol{y}^{T}) \label{eq:def_bilinear}\text{,}
\end{align}
where $\boldsymbol{z} \in \mathbb{R}^{D}$ is the output vector, 
and function $Vec(\ )$ vectorizes a matrix. 
$\boldsymbol{x} \in \mathbb{R}^{C}$ and $\boldsymbol{y} \in 
\mathbb{R}^{C'}$ denotes two input vectors each containing 
$C$ channels. $\boldsymbol{W} \in \mathbb{R}^{D 
\times CC'}$ is the learnable parameters.

As the number of parameters of this naive bilinear module 
is too large for widely usage 
\cite{Gao2016CompactBP,Kim2017,Yu2017MultimodalFB}, 
we factorize each element $w_{kij}$ in weight $\boldsymbol{W} \in 
\mathbb{R}^{D\times C\times C'}$ by three smaller matrices: 
%\begin{align}
$
w_{kij} = 
    \sum_{r=1}^{R}u_{kr}a_{ir}b_{jr} \label{eq:approx}\text{,}
$
%\end{align}
where $(u_{kr}) = \boldsymbol{u} \in 
\mathbb{R}^{D \times R}$, $(a_{ir}) = \boldsymbol{a} \in 
\mathbb{R}^{C \times R}$, 
$(b_{jr}) = \boldsymbol{b} \in 
\mathbb{R}^{C' \times R}$ are factorized parameters. 
Then the General Approximated Bilinear Module ({\bf ABM-G}, 
Fig. \ref{fig:ABMs} (a)) can be defined as: 
\begin{align}
    \boldsymbol{z} &= ABM_{g}(\boldsymbol{x}, \boldsymbol{y}) \\
                   &= \boldsymbol{u}\cdot (
                        \boldsymbol{a}^{T}\boldsymbol{x} \circ 
                        \boldsymbol{b}^{T}\boldsymbol{y}) \label{eq:abm_g}
                        \text{,}
\end{align}
where $\circ$ denotes element-wise product. 
There are many variants of this form exploited in various 
applications \cite{Kim2017,Yu2017MultimodalFB,
Lin2015BilinearCM,Wang2018SMART}, and all of them can 
be derived from this general form by substituting some specific elements.
%Though the derived form is 
%similar to \cite{Kim2017,Yu2017MultimodalFB}, their approximation factorized 
%the weights by two sub-matrices while ours by three. We provide a complete 
%derivation in the appendix.

{\bf Relation to Two-Layer MLP.} For Eq. \ref{eq:abm_g}, if we fix 
$\boldsymbol{b}^{T}\boldsymbol{y} = \boldsymbol{1}$ and add 
a nonlinear layer in the middle \cite{Kim2017}, the ABM-G 
becomes a two-layer MLP. 
%: $\boldsymbol{z} = \boldsymbol{u}\cdot 
%(\boldsymbol{a}^{T}\boldsymbol{x})$. 
From this viewpoint, a two-layer 
MLP can be seen as a constrained approximation of the bilinear module, 
whose bilinear weights are factorized in a constrained way: 
$
w_{kij} = 
    \sum_{r=1}^{R}u_{kr}a_{ir}b_{jr} \label{eq:cons_approx}\text{,}
$
where 
$
\sum_{j=1}^{C'}b_{jr}y_{j} = 1 \label{eq:cons_approx2}\text{,}
$
with an additional activation layer for keeping nonlinearity. 
%If $\boldsymbol{y}=\boldsymbol{x}$, this constraint scales a 
%bilinear transformation of $\boldsymbol{x}$ down to a linear transformation; 
%if $\boldsymbol{y} \neq \boldsymbol{x}$, 
This constraint just ignores 
information from $\boldsymbol{y}$ thus no bilinear features are learned. 
Because of this negative effect, we refer this branch outputting 
$\boldsymbol{1}$ as constrained-branch. 
%which scales an ABM-G down to be a 
%two-layer MLP for doing only linear transformations. 
%If we free a two-layer MLP from this constraint, we can get a more 
%discrimitive module that learns a pair-wise 
%relation feature similar to bilinear operation. 
Reversely, if we free a two-layer MLP from this constraint by making the 
weights in the constrained-branch tunable, then the freed 
MLP, which is now ABM-G with an additional nonlinearity layer, 
can learn a bilinear feature 
rather than the original linear feature, 
and we name this tunable branch \emph{auxiliary-branch}. 
This transformation enables a pathway to enhance the 
traditional two-layer MLPs to be more discriminative, 
which is also the key technique how we 
implant the ABMs into CNNs' intermediate 
layers (Sec. \ref{sec:implant}), while reusing the pretrained 
weights.

\begin{figure*}[t]
\begin{center}
   \includegraphics[width=\linewidth]{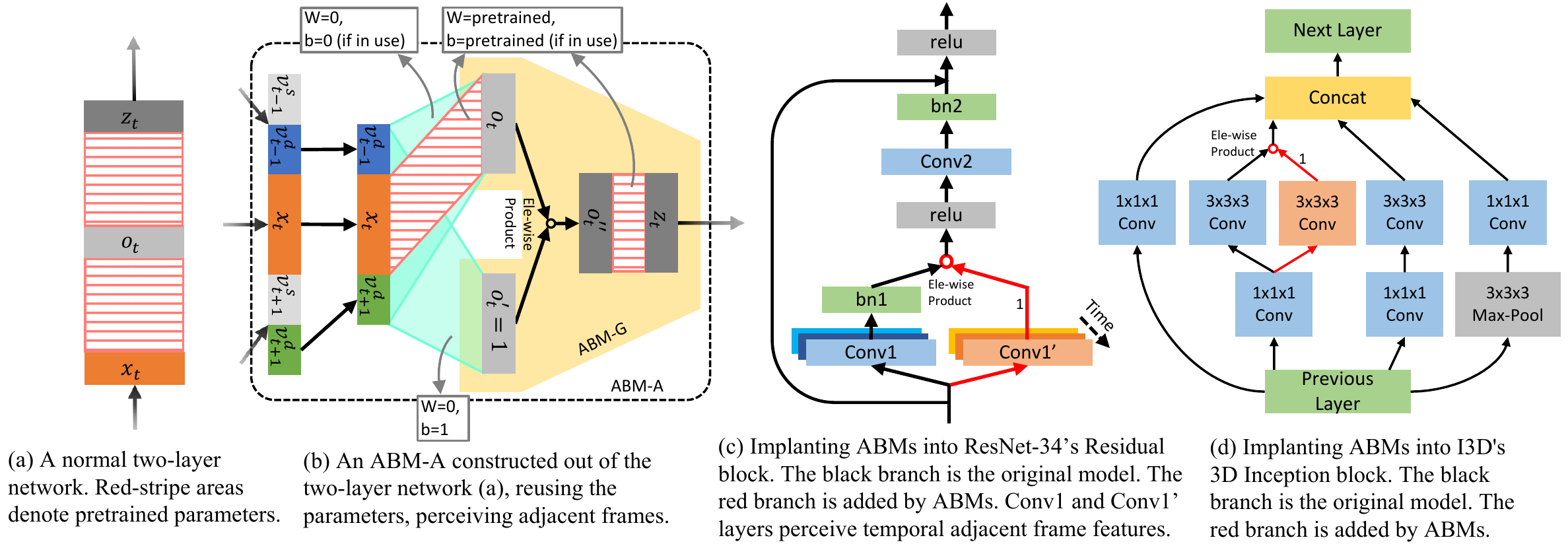}\vspace{-2.0ex}
\end{center}
    \caption{(a), (b): Structure details in ABM-A module. 
    Red-stripe areas are 
    pretrained parameters in the original CNN, 
    while cyan areas are newly initialized. 
    (c), (d): Network instantiations. 
    Red arrows are \emph{auxiliary-branches} with tunable weights 
    used to construct ABMs.}
\label{fig:mixed}\vspace{-2.0ex}
\end{figure*}

\subsection{Temporal ABMs}
Unlike other bilinear applications where inputs 
usually comes in duel forms, e.g. two branches 
\cite{lin2015bilinear,Gao2016CompactBP}, 
two modalities \cite{Fukui2016MultimodalCB,Kim2017,
Yu2017MultimodalFB,Wang2017SpatiotemporalPN}, it is not 
very straightforward to apply bilinear modules to temporal problems. 
We consider several variants for temporal modeling. The frame features along 
time are denoted as $\{\boldsymbol{x}_{1}, \boldsymbol{x}_{2}, .., 
\boldsymbol{x}_{t}, ..\}$.

{\bf ABM-S.} First we consider simply feeding 
adjacent frame features to ABM-G's two entries separately:
%\begin{align}
$
    \boldsymbol{z} = ABM_{g}(\boldsymbol{x}_{t}, 
        \boldsymbol{x}_{t+1}) \label{eq:cross_frame} \text{.}
$
%\end{align}
This is the most straightforward way and easy to implement. 
We name it ABM-S (see Fig. \ref{fig:ABMs} (b)). 
The potential drawbacks of this variant are: 
1. its temporal receptive field is limited since it can only 
perceive two frames at once; 2. it lacks self-bilinear capability 
which is shown effective for classification in some cases 
\cite{Kong2017LowRankBP,Li2017FactorizedBM,Diba2017DeepTL}.
We propose ABM-C to solve these problems.

{\bf ABM-C.} We consider a second way to feed a 
concatenation of multiple frames into both 
ABM entries:
\begin{align}
    \boldsymbol{z} &= ABM_{c}(\boldsymbol{x}_{
        t-\lfloor m/2 \rfloor},..,
        \boldsymbol{x}_{t}, 
        ..,\boldsymbol{x}_{t+\lfloor m/2 \rfloor}) \\
    &= ABM_{g}(\boldsymbol{x}'_{t}, 
        \boldsymbol{x}'_{t}) \label{eq:along_frame} \text{,} \\
    \boldsymbol{x}'_{t} &= Concat(\boldsymbol{x}_{
        t-\lfloor m/2 \rfloor},..,
    \boldsymbol{x}_{t}, 
    ..,\boldsymbol{x}_{t+\lfloor m/2 \rfloor}) \text{,}
\end{align}
where $m$ denotes the number of concatenated frames. 
We name this module ABM-C (see Fig. \ref{fig:ABMs} (c)). 
This variant can perceive more frames at once and it is 
similar to naive convolution so it is easier to work with existing CNNs. 
In this paper, we fix $m=3$. We show that ABM-C is 
more effective than ABM-S in the 
experiments (Sec. \ref{sec:ablation}).
A potential problem of this module is its massive parameters since 
its parameter number grows linearly with the perceived frames. 
To diminish this problem, we propose ABM-A. 

{\bf ABM-A.} We consider an intrinsic property of 
videos: repetition.  For most time, adjacent frames come 
with duplicated visual information, and 
the dynamics in them that defines the motion of a video is very 
subtle and fine-grained. Therefore we propose to divide a single 
frame descriptor into two parts: 
\begin{align}
    \boldsymbol{x}_{t} = Concat(\boldsymbol{v}^{s}_{t}, 
    \boldsymbol{v}^{d}_{t}) \label{eq:repetition} \text{,}
\end{align}
where $\boldsymbol{v}^{s}_{t}$ is the part containing 
information that looks 
static to the adjacent frames, and $\boldsymbol{v}^{d}_{t}$ containing 
dynamic information. In other words, we hypothesize that the static part 
is shared in the short local snippet and it 
mainly contains static visual information: 
$\boldsymbol{v}^{s}_{t-1} \approx \boldsymbol{v}^{s}_{t} \approx 
\boldsymbol{v}^{s}_{t+1}$ 
\footnote{Here we only show a snippet of 3 frames.}, while the dynamic 
part is more discriminative for temporal modeling.

Based on this intuition, we define the Adjacent 
Approximated Bilinear Module (ABM-A): 
\begin{align}
    \boldsymbol{z}_{t} &= ABM_{a}(\boldsymbol{x}_{t-1}, 
    \boldsymbol{x}_{t}, \boldsymbol{x}_{t+1}) \\
    &= ABM_{g}(\boldsymbol{x}''_{t}, \boldsymbol{x}''_{t}) 
        \label{eq:abm_a} \text{,} \\
    %\boldsymbol{x}''_{t} &= Concat(\boldsymbol{v}^{s}_{t}, 
        %\boldsymbol{v}^{d}_{t}, \boldsymbol{v}^{d}_{t-1}, 
        %\boldsymbol{v}^{d}_{t+1}) \text{.}
    \boldsymbol{x}''_{t} &= Concat(\boldsymbol{x}_{t}, 
        \boldsymbol{v}^{d}_{t-1}, 
        \boldsymbol{v}^{d}_{t+1}) \text{.}
\end{align}
See Fig. \ref{fig:ABMs} (d) for an illustration. 
Since $|\boldsymbol{x}_{t}| = |\boldsymbol{v}^{s}_{t}| + 
|\boldsymbol{v}^{d}_{t}|$, we can define 
$\beta = \frac{|\boldsymbol{v}^{d}_{t}|}{|\boldsymbol{x}_{t}|} \in [0,1]$ 
as a hyper-parameter. When $\beta = 0$, the module becomes purely frame-level 
and conducts no temporal modeling; when $\beta = 1$, 
the module is equal to ABM-C for full temporal modeling. 
We investigate 
$\beta = 
\frac{1}{2}$ and $\frac{1}{4}$ in the experiments 
(Sec. \ref{sec:ablation}). 

%{\bf Relation to TSM.} Recently Lin \etal \cite{LinTSM} introduced 
%TSM module which shifts features in temporal domain to do temporal 
%modeling. 
%Though both our and their modules 
%incorporate adjacent partial features 
%into the current frame, TSM discards part of the current frame feature while 
%our module does not. Also in TSM, shifted features 
%are of different parts while in ours they are the same. 
%Besides, our modules are to approximate a bilinear operation 
%while TSM is a purely linear module.

\subsection{Exploiting Deep Architectures} \label{sec:implant}
%We are committed to develop a general module which can not only 
%work in a plug-and-play manner, but also enjoy the pretraining of 
%deep architectures (\eg on ImageNet or Kinetics). 
We are committed to develop our modules to be general and flexible 
enough so that it not only can work in a plug-and-play manner, 
but also can enjoy the pretraining of deep architectures. 
%In this subsection we introduce two ways to combine our ABM modules with 
%deep CNNs while keeping both the merits.
In this subsection we introduce how our ABM modules can work with 
existing deep CNNs, while reusing the pretrained parameters.

{\bf On Top of CNNs.}
The most straightforward way is to put our modules on top of 
deep CNNs so that the backbone is not interfered and all pretrained 
parameters could be intact. 
We stack multiple layers of ABM modules to increase the temporal 
receptive fields and capture more complex temporal nonlinearity. 
This model conducts all temporal modeling at the high-semantic level 
which can lead to a very high-speed inference. 
In this case, ABM parameters can be 
initialized randomly which is easy to implement. 
We compare this implementation with 
various post-CNN temporal pooling methods 
\cite{Diba2017DeepTL,Wang2016TemporalSN,Zhou2017TemporalRR} 
using a same backbone network with same configurations 
to fairly show our models' effectiveness. 
However this implementation has some problems: 1. since the ABMs are all 
initialized randomly and contain element-wise multiplications, it could not 
go too deep or may lead to convergence difficulty and slow down 
the training; 2. because of the first problem, it could not have a 
large temporal receptive field, leading to limited temporal modeling 
capability. 
%\begin{figure}[b]
%\begin{center}
   %\includegraphics[width=0.9\linewidth]{imgs/ABMs_init}
%\end{center}
    %\caption{Structure and initialization details in ABM-A module. 
    %Red-stripe areas 
    %are initialized with pretrained parameters of two cascaded convolutional 
    %layers in the original CNN, while cyan areas are initialized as shown 
    %in the figure.}
%\label{fig:mixed}
%\end{figure}

{\bf Implanted into CNNs.} Additionally, we consider a more 
flexible way: implanting ABMs into deep architectures's intermediate layers. 
As we show in Sec. \ref{sec:abm_define} that MLPs could 
be seen as a constrained approximation of bilinear operations, we could 
also reversely transform two-layer convolutions of CNNs into ABMs 
by constructing an \emph{auxiliary-branch} with tunable weights.
%as 
%we know convolutions are linear transformations that happens in a 
%kernel-size scale. 

Let's assume we want to build an ABM-A module out of 
two convolutional layers (Fig. \ref{fig:mixed} (a)), 
which should be quite common in deep CNN architectures. 
%For the first convolution $Conv1$ (see $\boldsymbol{x}_{t} \rightarrow 
%\boldsymbol{o}_{t}$ in Fig. \ref{fig:mixed} (a) and (b)), 
Firstly we retain the first convolution $Conv1$ 
(see $\boldsymbol{x}_{t} \rightarrow 
\boldsymbol{o}_{t}$ in Fig. \ref{fig:mixed} (a) and (b)). 
Secondly we construct its sibling operation 
$Conv2$ (see $\boldsymbol{x}_{t} \rightarrow 
\boldsymbol{o}'_{t}$ in Fig. \ref{fig:mixed} (b)), 
containing the same input and output dimensions as $Conv1$. 
Then we initialize all its weights to be $\boldsymbol{0}$, 
and the corresponding bias to be $\boldsymbol{1}$ so 
that initially whatever the input is, the output of 
$Conv2$ is $\boldsymbol{1}$. 
By taking the above two steps, 
we have manually constructed the \emph{auxiliary-branch} 
($\boldsymbol{b}^{T}\boldsymbol{y} = \boldsymbol{1}$, see 
Sec. \ref{sec:abm_define} for explanation). 
This guarantees that: $Conv1(x)\circ Conv2(y) = Conv1(x)$, 
meaning the original pathway in the CNN is intact, 
therefore the pretrained parameters are still valid. 
By freeing the weights in the \emph{auxiliary-branch}, 
we get an ABM-G whose initial 
power is the same as the original two-layer network.

If the original CNN is a 2D network (\eg pretrained on ImageNet), 
we also need to adapt it for temporal modeling. 
In this case, the input of 
$Conv1$ and $Conv2$ is expanded to 
perceive surrounding frame features. 
How the expansion happens depends on the 
type of ABM in used here, \eg for ABM-A/C, we need to 
incorporate the dynamic feature parts of adjacent frames; 
for ABM-S, two branches take two neighbored frames as inputs separately. 
However only the weights corresponding to the current frame are initialized 
with pretrained parameters while others are set to be zeros (cyan 
areas in Fig. \ref{fig:mixed} (b)). 
This modification also 
does not change the behavior of the origional 2D network so the 
pretrained parameters are still valid (red-stripe areas in 
Fig. \ref{fig:mixed} (b) denote the 
original CNN pathway with pretrained parameters). 
Together with the second 
convolution layer (see $\boldsymbol{o}''_{t} 
\rightarrow \boldsymbol{z}_{t}$ in Fig. \ref{fig:mixed} (b), 
which is also $\boldsymbol{o}_{t} \rightarrow \boldsymbol{z}_{t}$ 
in Fig. \ref{fig:mixed} (a)), 
we built an ABM-A out of pretrained two-layer convolutions 
with all parameters preserved, based on the idea that 
freeing the \emph{auxiliary-branch} to be tunable. 
If the origional CNN is already 3D, we can 
construct ABM-C modules by just adding 
the \emph{auxiliary-branch} with 
initialization of $\boldsymbol{W}=\boldsymbol{0}$, 
$\boldsymbol{b}=\boldsymbol{1}$. 
%By making the parameters tunable, an ABM-C module is built.
By implanting ABMs into CNNs with careful initializations, 
the ABMs can be stacked very deeply with 
temporal receptive fields becoming very large. 
%We show this results in faster 
%convergence and better performance. 

%\begin{figure}[t]
%\begin{center}
   %\includegraphics[width=0.8\linewidth]{imgs/res34_abm}
%\end{center}
    %\caption{Implanting ABMs into Residual Blocks. The black branch 
    %is the original model. The red branch is 
    %added by ABMs. The multi-layer $conv1$ and $conv1'$ means taking 
    %temporal adjacent frame features as input.}
%\label{fig:res34_abm}
%\end{figure}

%\begin{figure}[t]
%\begin{center}
   %\includegraphics[width=0.8\linewidth]{imgs/i3d_abm}
%\end{center}
    %\caption{Implanting ABMs into I3D's 3D Inception block. The black branch 
    %is the original model. The red branch is 
    %added by ABMs.}
%\label{fig:i3d_abm}
%\end{figure}

{\bf Network Instantiations.} We instantiate our proposed modules 
with two architectures: 2D-ResNet-34 \cite{He2016DeepRL} and 
I3D \cite{Carreira2017QuoVA}. The 2D-ResNet-34 backbone 
is used to show the true power of 
ABMs for temporal modeling. This backbone is pure 2D and 
pretrained on ImageNet dataset \cite{imagenet_cvpr09} 
for image classification without 
any prior knowledge about temporal information. 
The I3D backbone is used to show the complementarity between our ABMs and 
the state-of-the-art 3D architecture. 
It is pretrained on Kinetics dataset \cite{kinetics_dataset} 
for action recognition. 

For 2D-ResNet-34, we implant ABMs into each non-down-sampling 
residual blocks for block-layer 2, 3, and 4. 
In each block, there are two convolutional 
layers which is perfect for ABMs' construction. Following last section, 
we can add a tunable \emph{auxiliary-branch} and expand the temporal 
receptive field to build ABM modules (see 
Fig. \ref{fig:mixed} (c)). We add a kernel-2 stride-2 temporal maxpooling 
layer after block-layer 2 for more efficient computation. 

For I3D, we build ABMs aside the larger $3 \times 3 \times 3$ convolution 
in each Inception block after layer-3c (see Fig. \ref{fig:mixed} (d)). 
Though there is no appended layer 
to form a complete ABM in a single Inception block, by taking into account 
the next Inception block the ABM is still complete. 

\subsection{Snippet Sampling} \label{sec:sampling}
In \cite{Zhou2017TemporalRR,ECO_eccv18,LinTSM}, 
sparse sampling is used for efficient 
video processing. Specifically, they divide each video into $N$ segments 
and sample a single frame from each. We generally follow this strategy, but 
also argue that only sampling a single frame per segment discards too much 
useful short-term information in consecutive frames. 
Instead, we borrow 
a strategy from optical flow sampling 
\cite{Simonyan2014TwoStreamCN}, which each time samples a short snippet 
(containing $K$ frames) rather than only a single frame. 
To keep efficiency of sparse sampling, 
only the weights in the first convolutional layer are duplicated to perceive 
the snippet, while rest of the network still feels it is processing 
$N$ single frames. In this paper, we choose $N=8$ or $16$, and fix $K=3$ 
after ablation study. Snippet sampling is represented by 
$N \times K$ in tables.

\subsection{Implementation Details} \label{sec:implementation}
{\bf Training.} For inputs, we randomly sample a snippet in each segment 
for 2D-ResNet-34 models, and densely 
sample 64 frames for I3D models. 
The input frames are scaled to $256 \times 256$ and randomly 
cropped to $224 \times 224$. For Something-Something v1 and v2 
datasets, we use 2D-ResNet-34 backbone pretrained on ImageNet to 
show the pure effectiveness of our modules' temporal modeling; for 
other datasets, we use I3D backbone pretrained on Kinetics. We 
train all models using SGD with momentum of 0.9. 
All models are trained or fine-tuned with 0.001 initial learning 
rate and decayed by 10 twice. For 2D-ResNet-34-top models, 
lr decays at epoch 30 and 40 (total 50 epochs). 
For 2D-ResNet-34-implanted models, lr decays at epoch 15 and 20 
(total 25 epochs). Because of limited GPU resources, 
I3D-based models are first trained on Kinetics for 8 epochs, 
and fine-tuned for 20 epochs on smaller datasets. The lr decays every 
8 epochs during fine-tuning. All experiments are conducted on 4 GeForce 
GTX TITAN X gpus.
% 4 GeForce GTX TITAN X

{\bf Testing.} During testing of 2D-ResNet-34-based models, 
we sample the center snippet for 
each segment to do the inference. We also introduce a new testing 
protocol by shifting the snippets so that more frames are used in a 
segment. We define shifting-time $ST$, so shifted samples of a video will 
be used for inference, and the output of a video will be the averaged 
output of each shifted sample. To calculate the shifting-offset, we 
divide each segment by $ST$. In this paper, we fix $ST=3$, and 
we will specify if shifting inference is used in the experiments. 
For I3D-based models, center 150 frames are used for inference. 
During validation and testing, the video frames are scaled 
to $224 \times 224$ then fed to models. No other 
cropping strategies are used.

%------------------------------------------------------------------------
\section{Experiments}
We perform comprehensive ablation studies on Something-v1 
dataset \cite{Goyal2017TheS}. 
%to show the effectiveness of proposed 
%model variants and techniques. 
Then we compare our models with 
state-of-the-art methods on various datasets, while showing our models' 
generality to optical flow modality. 
Efficiency analysis and visualization are also provided. 

\subsection{Datasets}
Something-Something v1 \cite{Goyal2017TheS} and 
v2 \cite{Mahdisoltani2018FinegrainedVC} are crowd-sourced 
datasets focusing on temporal modeling, containing fine-grained 
human motions and human-object interactions. There are 
108,499 videos in v1 and 220,847 videos in v2, with 174 categories 
in each dataset. Besides, other YouTube-like datasets, 
Kinetics \cite{kinetics_dataset}, 
UCF101 \cite{Soomro2012UCF101AD}, and HMDB51 \cite{Kuehne2011HMDBAL}, 
are also used to validate our models on action recognition. 
These three datasets contain more static-recognizable 
categories thus spatial modeling could be as crucial as temporal modeling. 
The latter two sets are small-scale so Kinetics-pretraining is used.

%{\bf Something-Something-v1.} This dataset contains 174 classes of 
%108,499 videos, 86,017 
%for training, 11,522 for validation, and 10,960 for testing. 
%Because the interacted 
%objects can vary drastically for a single class, it is very difficult to 
%correctly classify a video by single frame.

%{\bf Something-Something-v2.} It is a large-scale fine-grained action 
%recognition dataset containing 220,847 videos, 168,913 
%for training, 24,777 for validation, and 27,157 for testing. It has the 
%same classes and similar properties as the last one, 
%but contains less label noise.

%{\bf Kinetics.} This is a large-scale traditional action recognition 
%dataset containing ~246k training videos and 20k validation videos, 
%which involves 400 categories. Unlike the two above crowdsourced 
%datasets, this dataset is labeled more visually thus backbone network 
%could be the key for high performance.

%{\bf UCF101.} This is a widely benchmarked middle-scale action recognition 
%dataset of Kinetics type, containing 13,320 YouTube 
%clips spanning over 101 categories, 
%and has 3 training and validation splits. 

%{\bf HMDB51.} This set contains 6,766 clips of 51 action classes with 3 
%training and validation splits. The reported results for both 
%UCF101 and HMDB51 are averaged over the 3 official splits.

\subsection{Ablation Study on Something-v1} \label{sec:ablation}
%We conduct ablation studies on Something-v1. 

%\begin{table}
%\begin{center}
%%\footnotesize
%\small
%\begin{tabular}{l|l|c|c|c|c}
%\hline
    %\# & Model & Top-1 & Top-5 & Para\# & rVPS\\
%\hline\hline
    %\multirow{3}{*}{1} & Avg Pooling & 18.09 & 43.67 
        %& 21.1M & 1.00$\times$ \\
    %& TRN \cite{Zhou2017TemporalRR} & 31.68 & 60.61 & 21.4M & 0.93$\times$ \\
    %& CBP \cite{Gao2016CompactBP,Diba2017DeepTL} 
        %& 34.40 & 60.70 & 22.5M & 0.14$\times$ \\
    %%& TBA \cite{Girdhar2017AttentionalPF} & 28.22 & 54.85 
        %%& 21.2M & 0.92$\times$ \\
%\hline
%\hline
    %\multirow{4}{*}{2} & ABM-S-\emph{top} L=1 & 29.94 & 57.83 
        %& 21.7M & 0.96$\times$ \\
    %& ABM-C-\emph{top} L=1 & {\bf 35.49} & {\bf 64.11} & 25.4M & 0.92$\times$ \\
    %& ABM-S-\emph{top} L=3 & 30.31 
        %& 57.56 & 22.8M & 0.92$\times$ \\
    %& ABM-C-\emph{top} L=3 & {\bf 38.32} & {\bf 66.15} & 24.3M & 0.91$\times$ \\
%\hline
%\end{tabular}
%\end{center}
    %\caption{Results of \emph{On Top of CNNs} implementation and 
    %some baseline models on Something-v1 dataset. 
    %The baselines are reproduced by 
    %ourselves using 2D-ResNet-34 backbone for a fair comparison. 
    %rVPS (higher is better) is the relative video per second compared to 
    %Avg Pooling model (actual speed 228.6 VPS). 
%}
%\label{table:post_cnn}
%\end{table}

\begin{table}
\begin{center}
%\footnotesize
%\small
\begin{tabular}{l c c c r}
\hline\hline
    Model & \#Frame & Top-1 & Top-5 & VPS \\
\hline\hline
    Avg Pooling & 8 & 18.09 & 43.67 & 85.81 \\
    TRN \cite{Zhou2017TemporalRR} & 8 & 31.68 & 60.61 & 84.22 \\
    CBP \cite{Gao2016CompactBP,Diba2017DeepTL} 
        & 8 & 34.40 & 60.70 & 65.50 \\
\hline
    ABM-S-\emph{top} L=1 & 8 & 29.94 & 57.83 & 84.84 \\
    ABM-S-\emph{top} L=3 & 8 & 30.31 & 57.56 & 80.56 \\
    ABM-C-\emph{top} L=1 & 8 & 35.49 & 64.11 & 84.82 \\
    ABM-C-\emph{top} L=3 & 8 & 38.32 & 66.15 & 83.66 \\
    ABM-C-\emph{top} L=3 & 8$\times$2 & 41.01 & 68.46 & 82.07 \\
    ABM-C-\emph{top} L=3 & 8$\times$3 & {\bf 42.35} & {\bf 71.82} & 80.38 \\
\hline
\end{tabular}
\end{center}
    \caption{Results of \emph{On Top of CNNs} implementation and 
    some other temporal models on Something-v1 dataset. 
    %The baselines are reproduced by 
    %ourselves using 2D-ResNet-34 backbone for a fair comparison. 
    L denotes the number of ABM-C layers. 
    %K denotes sampled snippet length (see 
    %\emph{Sampling Strategy} in Sec. \ref{sec:sampling}). 
    VPS means \emph{video per second}. 
}
\label{table:post_cnn}
\end{table}

\begin{table}
%\small
\begin{center}
\begin{tabular} {l c c c c}
\hline\hline
    Model & \#Frame & Top-1 & Top-5 & VPS \\
        %& forward time & backward time \\
\hline\hline
    ABM-C-\emph{top} & 8$\times$3 & 42.35 & 71.82 & {\bf 80.38} \\
    ABM-C-\emph{in} & 8$\times$3 & 44.14 & 74.16 & 28.02 \\
\hline
    ABM-A-\emph{in} $\beta$=1/4 & 16$\times$3 & 44.89 & 74.62 & 40.19 \\
    ABM-A-\emph{in} $\beta$=1/2 & 16$\times$3 & 45.67 & {\bf 74.80} & 36.65 \\
    ABM-A-\emph{in} $\beta$=1 (C) & 16$\times$3 & {\bf 46.08} & 74.32 & 20.12 \\
%\hline
    %ABM-C-\emph{top}-\emph{sh} & 8$\times$3 & * & * & * \\
    %ABM-C-\emph{in}-\emph{sh} & 8$\times$3 & * & * & * \\
    %ABM-C-\emph{in}-\emph{sh} & 16$\times$3 & {\bf 46.81} & * & * \\
\hline
\end{tabular}
\end{center}
    \caption{Comparison among different ABM variants. \#Frame is shown in 
    $N \times K$ to indicate the snippet sampling.}
    \label{table:work_with_cnn}
\end{table}

\begin{figure}[t]
\begin{center}
   \includegraphics[width=\linewidth]{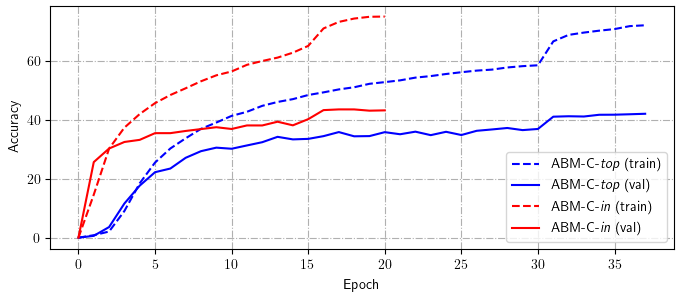}\vspace{-1.5ex}
\end{center}
    \caption{Training curves of 
    ABM-C-\emph{top} and ABM-C-\emph{in} models.}
\label{fig:learning_curve}\vspace{-1.5ex}
\end{figure}

\begin{table}
%\small
\begin{center}
\begin{tabular} {l c c c}
\hline\hline
    Inference & ABM-A $\beta$=1/4 & ABM-A $\beta$=1/2 & ABM-C \\
        %& forward time & backward time \\
\hline\hline
    w/o Shift & 44.89 & 45.67 & 46.08 \\
    w/ Shift & {\bf 45.56} & {\bf 46.16} & {\bf 46.81} \\
\hline
\end{tabular}
\end{center}
    \caption{Shifting Inference. The shifting-time is fixed to 3.}
    \label{table:shifting}\vspace{-1.5ex}
\end{table}

{\bf Top-of-CNN Effectivsness.} We first conduct experiments with 
\emph{On Top of CNNs} structure using 8-frame sampling. 
No shifting inference is used. 
Results are shown in Table \ref{table:post_cnn}. 
For comparison, we reimplement several existing 
post-CNN temporal models: 1. Avg Pooling or naive version of TSN 
\cite{Wang2016TemporalSN}; 2. TRN \cite{Zhou2017TemporalRR}; 
3. Compact Bilinear Pooling \cite{Gao2016CompactBP,Diba2017DeepTL}. 
To calculate VPS (\emph{video per second}), we do validation 
on a single GeForce GTX TITAN X 
with fixed batch size of 8, and discard the first 200 iterations to 
calculate a stable VPS. We can see the 
ABM-S-\emph{top} models, which perceive 
neighbored frames seperately by two branchs, are not 
very effective even we increase the ABM-S layers to 3. On the other hand, 
ABM-C-\emph{top} models, 
whose both branches perceive concatenated frames, are 
very effective and can enjoy lots of gain from more layers. 
%Therefore we discard the ABM-S module in the future experiments. 
%We use ABM-C as our default module in the future experiments. 

%{\bf Does Snippet Sampling Work?} In Table \ref{table:work_with_cnn}, 
%we can see that snippet sampling is a very effective way 
%to boost performance (around $4\%$ by increasing K from 1 to 3). 
%We fix $\text{K}=3$ in the future experiments. 

%==== State of the art on Something
\begin{table*}
\begin{center}
\small
%\footnotesize
\begin{tabular}{l|c|c|c|c|c|c|c|c}
%\begin{tabularx}{\textwidth}{X|c|c|c|c|c|c|c}
\hline\hline
    Model & Pretrain & Modality & \#Frame & Backbone & 
    v1-Val & v1-Test & v2-Val & v2-Test \\
\hline\hline
    Multi-Scale TRN \cite{Zhou2017TemporalRR} & ImgN & RGB & 
        8 & BN-Inception & 34.44 & 33.60 & 48.80 & 50.85 \\
    %2D-VGG-LSTM \cite{Mahdisoltani2018FinegrainedVC} & ImgN & RGB & 
        %48 & 2D-VGG & - & 31.69 & 40.20 & 38.83 \\
    3D-VGG-LSTM \cite{Mahdisoltani2018FinegrainedVC} & ImgN & RGB & 
        48 & 3D-VGG & - & - & 51.96 & 51.15 \\
    ECO-Lite$_{En}$ \cite{ECO_eccv18} & Kin & RGB & 
        92 & 2D-Inc+3D-Res & 46.4 & 42.3 & - & - \\
    NL I3D \cite{Wang_videogcnECCV2018} & Kin & RGB & 
        64 & 3D-ResNet-50 & 44.4 & - & - & - \\
    NL I3D+GCN \cite{Wang_videogcnECCV2018} & Kin & RGB & 
        64 & 3D-ResNet-50 & 46.1 & {\bf 45.0} & - & - \\
    TSM \cite{LinTSM} & Kin & RGB & 
        16 & 2D-ResNet-50 & 44.8 & - & 58.7 & 59.9 \\
    TSM$_{En}$ \cite{LinTSM} & Kin & RGB & 
        24 & 2D-ResNet-50 & 46.8 & - & - & - \\
\hline
    ABM-C-\emph{in} & ImgN & RGB & 16 & 2D-ResNet-50 & 
        47.45 & - & - & - \\
    ABM-C-\emph{in} & ImgN & RGB & 16$\times$3 & 2D-ResNet-50 & 
        {\bf 49.83} & - & - & - \\
\hline
    ABM-A-\emph{in} \emph{$\beta$}=1/2 & ImgN & RGB & 16$\times$3 & 2D-ResNet-34 & 
        46.16 & - & - & - \\
    ABM-C-\emph{in} & ImgN & RGB & 16$\times$3 & 2D-ResNet-34 & 
        46.81 & - & {\bf 61.25} & {\bf 60.13} \\
    ABM-AC-\emph{in}$_{En}$ & ImgN & RGB & 32$\times$3 & 
        2D-ResNet-34 & {\bf49.02} & 42.66 & - & - \\
\hline\hline
    %Non-local \cite{NonLocal2018} & ImgN+Kin & RGB & 
        %TSN Dense & 16 & 3D-ResNet-50 & - & 45.04 \\
%\hline
    Multi-Scale TRN \cite{Zhou2017TemporalRR} & ImgN & RGB+Flow & 
        8+8 & BN-Inception & 42.01 & 40.71 & 55.52 & 56.24 \\
    ECO-Lite$_{En}$ \cite{ECO_eccv18} & Kin & RGB+Flow & 
        92+92 & 2D-Inc+3D-Res & 49.5 & 43.9 & - & - \\
    TSM \cite{LinTSM} & Kin & RGB+Flow & 
        16+8 & 2D-ResNet-50 & 49.6 & {\bf 46.1} & 63.5 &  {\bf 63.7} \\
\hline
    %HV Model L=3 & ImgN & RGB & 8 & 2D ResNet-34 & 
        %38.32 & - & - & - \\
    %HV Model L=3,f=3 & ImgN & Flow & 8$\times$3 & 2D ResNet-34 & 
        %{\bf 30.34} & .. & {\bf 47.17} & .. \\
    %ABM-C-\emph{in} & ImgN & RGB+Flow & $($16$+$16$)\times$3 & 
        %2D-ResNet-34 & 50.09 & - & 63.36 & * \\
    ABM-C-\emph{in} & ImgN & RGB+Flow & $($16$+$16$)\times$3 & 
        2D-ResNet-34 & 50.09 & - & {\bf 63.90} & 62.18 \\
    ABM-AC-\emph{in}$_{En}$ & ImgN & RGB+Flow & $($32$+$16$)\times$3 & 
        2D-ResNet-34 & {\bf 51.77} & 45.66 & - & - \\
\hline
\end{tabular}
%\end{tabularx}
\end{center}
    \caption{State-of-the-art comparison on Something-v1 
    and v2 datasets. 
    %SS means \emph{Single-Scale} sampling, 
    %MS means \emph{Multi-Scale} sampling, 
    ${En}$ means an ensemble model, 
    ImgN means pretrained on \emph{ImageNet}, 
    and Kin means pretrained on \emph{Kinetics}. 
    $N \times K$ means snippet sampling (see Sec. \ref{sec:sampling}). 
    Grouped by input modalities.}
    \label{table:something_SOTA}\vspace{-1.5ex}
\end{table*}

{\bf Does Snippet Sampling Work?} 
We show the effect of snippet sampling by increasing 
snippet length from 1 to 3 on 
ABM-C-\emph{top} L=3. Results are in 
Table \ref{table:post_cnn} bottom. There is an 
obvious gain when snippet length is increased by just a small 
number, \eg 4\% boost from $\text{K}=1$ to $\text{K}=3$ while the 
inference time is almost not affected. 
This is because the additional calculation only 
comes from the first convolutional layer. 
However if a snippet is longer than 3, the data loading time will 
surpass the inference time, slowing down the training, 
so the snippet length is not further increased.
%However, we found the greatest bottleneck of increasing snippet length becomes 
%the data loading time, therefore we do not experiment with larger snippets, 
%when the training time starts to increase. 
%However, we still see a drop on 
%inference speed, preventing us from further increasing the 
%snippet length since it makes the model cumbersome. 

{\bf How to work with CNNs?} In Table \ref{table:work_with_cnn} first half, 
we compare \emph{On Top of CNNs} (denoted by \emph{top}) and 
\emph{Implanted into CNNs} (denoted by \emph{in}) frameworks. 
We see the \emph{top} model can run at a very high inference speed, while 
the \emph{in} model can achieve higher performance. But in 
Fig. \ref{fig:learning_curve} we can see the \emph{top} model converges 
much slower. This is due to the better initialization of \emph{in} 
models. 
Also there is a little training difficulty at the beginning of 
ABM-C-\emph{top}, which is more obvious when layers are over 4. 
%In fact, when ABM layers 
%of \emph{top} models are set over 4, 
%their convergence can be a lot slower, which limits the power of 
%\emph{top} models. 
Therefore we prefer to use ABMs by implanting them 
into CNNs, but the \emph{top} models are more useful when 
inference speed is a main concern. 

In Table \ref{table:work_with_cnn} second half, we compare ABM-A models 
and ABM-C models. An ABM-A becomes ABM-C when $\beta=1$. 
We can see $\beta$ controls the speed-accuracy tradeoff 
when it is shifting, but accuracy seems to be very 
similar between $\beta=1/2$ and $\beta=1$. We consider 
ABM-A with $\beta=1/2$ as a model with balanced performance 
and efficiency. 

{\bf Does Shifting Inference Work?} In Table \ref{table:shifting}, we 
see shifting inference can boost all model variants by around $0.7\%$. 
The shift-time is fixed to 3. 
We use this testing protocol in the state-of-the-art comparison.

\subsection{Comparison with State-of-the-Art}
We compare our ABM models 
with other state-of-the-art methods on Something-v1 and v2 datasets 
since these two datasets focus on temporal modeling.
Our ImageNet-pretrained ABM models can 
outperform most other methods, 
showing high effectiveness for temporal modeling.

{\bf RGB Models.} In the first half section of Table 
\ref{table:something_SOTA}, we compare state-of-the-art RGB models. 
The only models that have the same pretraining setting as ours are 
Multi-Scale TRN \cite{Zhou2017TemporalRR} 
and a baseline model 3D-VGG-LSTM \cite{Mahdisoltani2018FinegrainedVC}. 
Under this setting, the two methods can only achive very limited 
performance, outperformed by ours by about $10\%$ of accuracy on both sets. 
With the pretraining of Kinetics dataset, ECO-Lite$_{En}$ \cite{ECO_eccv18}, 
NL I3D+GCN \cite{Wang_videogcnECCV2018}, and TSM$_{En}$ \cite{LinTSM} can 
obtain competitive performance, using deeper backbone architectures. 
Besides their heavier pretraining, 
two of these methods are ensemble models and the other one reqires an 
MSCOCO pretrained Regional Proposal Network, which lead to 
an unfair comparison. However, despite the weaker configurations, 
our single model can still outperform all of them on the 
v1 validation set (acc 46.81\% with ResNet-34), and our ensemble model 
(ABM-C-\emph{in} + ABM-A-\emph{in} \emph{$\beta=1/2$})
can achieve an accuracy of 49.02\% using ResNet-34. 
Equipped with ResNet-50, our models can achieve 
47.45\% without snippet sampling or shifting inference 
(to show a fairer comparison), and 
49.83\% with these techniques, outperforming all 
methods under similar settings. 
On v2 set, our model can 
outperform all three models on validation (61.25\%) 
and testing sets (60.13\%). 
It even outperforms TSM which utilized 5-crop protocol while we are 
feeding $224\times 224$ single scaled frames to the model. 
%This is a strong evidence that our ABM is more 
%effective for temporal modeling than other temporal techniques.

{\bf Two-Stream Models.} We train an ABM-A-\emph{in} $\beta$=1/2 model using 
flow modality. This model achieves accuracy of $37.82\%$ on 
v1 and $53.85\%$ on v2 validation sets. 
We add its predictions to ABM-C-\emph{in} 
and ABM-AC-\emph{in}$_{En}$ by weight of 1:1 (second half in 
Table \ref{table:something_SOTA}). The two-stream models 
achieve $50.09\%$ and $51.77\%$ accuracy on v1 validation set, 
outperforming all other methods. 
Our model is a bit worse than TSM on both test sets. 
We believe the Kinetics-pretraining provides more generality, 
and their 5-crop protocol and deeper backbone are also very 
beneficial for better performance. 
As our models outperform most state-of-the-art methods under weaker 
settings, we conclude the ABMs are powerful modules for temporal modeling.

\begin{figure*}[t]
\begin{center}
   \includegraphics[width=\linewidth]{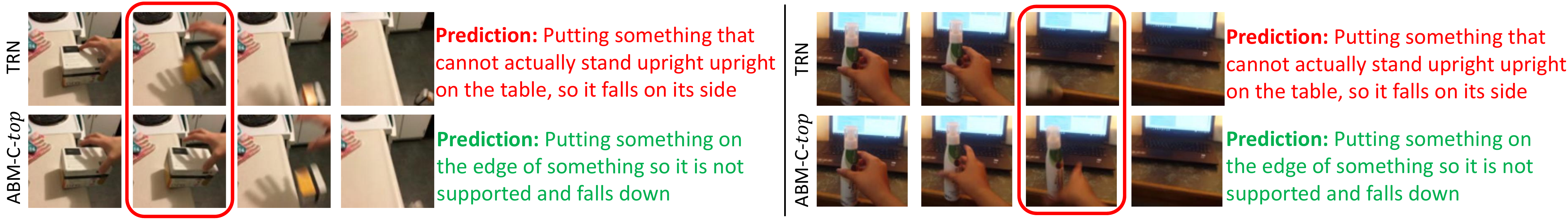}\vspace{-2.0ex}
\end{center}
    \caption{Keyframe visualization comparison between 
    TRN and ABM-C-\emph{top}. Wrong predictions are in red, 
    and correct predictions are in green. We highlight the 
    selected frames that visually differ most between two models, which 
    are the instant key moment in each video.}
\label{fig:keyframe}\vspace{-1.5ex}
\end{figure*}

\subsection{Efficiency Analysis}
We compare models' FLOPs and inference speed (by video per second) 
in Table \ref{table:efficiency} together with 
I3D \cite{Wang_videogcnECCV2018}, 
ECO \cite{ECO_eccv18}, and TSM \cite{LinTSM}. 
VPS of TSM is obtained by reimplementation using the official code. 
%VPS of TSM is inferred based on values in \cite{LinTSM} 
%because their code is not public available. 
%as their source code are public available. 
The measurement is conducted on a same single GeForce GTX TITAN X 
with batch size set to be 8 (2 for I3D because of memory limitation). 
As we see our heaviest model 
ABM-A-\emph{in} $\beta$=1 (is also ABM-C-\emph{in}) 
has 1/3 of FLOPs and 5 times faster than I3D 
while achieving higher accuracy. 
Compared with efficiency-oriented ECO, two of our models 
have smaller FLOPs, but can achieve higher accuracy and perceive 
more frames. Compared with TSM, four of our models inference faster, 
and three perform better.
%, thanks to our ABMs and snippet sampling. 

By comparing different ABM variants, we can clearly see 
the accuracy-speed tradeoff controlled by $\beta$: smaller $\beta$ 
results in lighter model and worse performance. However, the 
performance drop is not very huge, indicating representing 
frame features with separated static and dynamic parts is effective. 
As ABM-C-\emph{top} works significantly faster than \emph{in} models, 
it is an ideal model for online video understanding. 

\begin{table}
\begin{center}
%\footnotesize
\small
\begin{tabular}{l c c c r}
\hline\hline
    Model & \#Frame & FLOPs & VPS & Top-1 \\
\hline\hline
    I3D \cite{Wang_videogcnECCV2018} & 64 & 306G & 4.18 & 41.6 \\
    ECO \cite{ECO_eccv18} & 16 & 64G & 26.67 & 41.4 \\
    TSM \cite{LinTSM} & 16 & 65G & 20.10 & 44.8 \\
\hline
    ABM-A-\emph{in} $\beta$=1 (C) & 16$\times$3 & 106G & 20.12 & {\bf 46.08} \\ %& 61.95M \\
    ABM-A-\emph{in} $\beta$=1/2 & 16$\times$3 & 79G & 36.65 & 45.67 \\ %& 45.71M \\
    ABM-A-\emph{in} $\beta$=1/4 & 16$\times$3 & 67G & 40.19 & 44.89 \\ % & 37.60M \\
    ABM-C-\emph{in} & 8$\times$3 & 53G & 28.02 & 44.14 \\ %& 61.95M \\
    ABM-C-\emph{top} & 8$\times$3 & {\bf 32G} & {\bf 80.38} & 42.35 \\ %& 24.54M \\
\hline
\end{tabular}
\end{center}
    \caption{Efficiency comparison on Something-v1 dataset. 
    Models are put on a single GeForce GTX TITAN X with batch size 8 
    (2 for I3D) to calculate inference speed.}
\label{table:efficiency}
\end{table}

\begin{table}
\begin{center}
%\footnotesize
\small
\begin{tabular}{l c c c | c}
\hline\hline
    I3D \cite{Carreira2017QuoVA} & R(2+1)D \cite{Tran2017ACL} & 
    MF-Net \cite{Chen2018MultifiberNF} 
    %& S3D \cite{XieS3D} 
    & StNet \cite{DongliangHe2019AAAI} & ABM\\
\hline
    71.1 & 72.0 & {\bf 72.8} & 
    %72.2 & 
    71.38 & \bf{72.6} \\
\hline
\end{tabular}
\end{center}
    \caption{Comparison on Kinetics dataset.}
\label{table:kin_comp}
\end{table}

\begin{table}
\begin{center}
%\footnotesize
\small
%\begin{tabular}{l c c c c c}
\begin{tabular}{l c c c}
\hline\hline
    Model & Pre-Train & UCF101 & HMDB51 \\
\hline\hline
    C3D \cite{Tran2017ConvNetAS} & Sports-1M & 85.8 & 54.9 \\
    ARTNet + TSN \cite{Wang2018SMART} & Kinetics & 94.3 & 70.9 \\
    ECO$_{En}$ \cite{ECO_eccv18} & Kinetics & 94.8 & 72.4 \\
    I3D \cite{Carreira2017QuoVA} & Kinetics & {\bf 95.6} & {\bf 74.8} \\
    I3D (our impl) & Kinetics & 93.8 & 72.3 \\
\hline
    ABM-C-\emph{in} & Kinetics & 95.1 & 72.7 \\
\hline

%\hline\hline
    %Model & C3D \cite{Tran2017ConvNetAS} & ARTNet+TSN \cite{Wang2018SMART} & 
    %ECO$_{En}$ \cite{ECO_eccv18} & I3D \cite{Carreira2017QuoVA} & 
    %ABM-C-\emph{in} \\
%\hline\hline
    %UCF101 & 85.8 & 94.3 & 94.8 & {\bf 95.6} & 95.1 \\
    %HMDB51 & 54.9 & 70.9 & 72.4 & {\bf 74.8} & 72.7 \\
%\hline
\end{tabular}
\end{center}
    \caption{Comparison on UCF101 and HMDB51 datasets.}
\label{table:ucf101_comp}\vspace{-1.5ex}
\end{table}

%\begin{table}
%\begin{center}
%%\footnotesize
%\small
%\begin{tabular}{l l c c}
%\hline\hline
    %Model & Kinetics & UCF101 & HMDB51 \\
%\hline\hline
    %C3D \cite{Tran2017ConvNetAS} & 65.6 & 85.8 & 54.9 \\
    %ARTNet + TSN \cite{Wang2018SMART} & 70.7 & 94.3 & 70.9 \\
    %ECO$_{En}$ \cite{ECO_eccv18} & 70.0 & 94.8 & 72.4 \\
    %I3D \cite{Carreira2017QuoVA} & 71.1 & {\bf 95.6} & {\bf 74.8} \\
%\hline
    %ABM-C-\emph{in} & {\bf 72.6} & 95.1 & 72.7 \\
%\hline
%\end{tabular}
%\end{center}
    %\caption{State-of-the-art comparison on Kinetics, UCF101, and 
    %HMDB51 datasets by Top-1 accuracy using RGB inputs.}
%\label{table:ucf101_comp}
%\end{table}

\subsection{Results on Other Datasets}
We validate ABM-C-\emph{in} module equipped with I3D backbone 
on Kinetics, UCF101, and HMDB51 datasets using RGB inputs. 
Because of our limited GPU resources, we only train our model on Kinetics 
for 8 epochs, with initial learning rate of 0.001 and decay by 0.1 at epoch 4. 
Thanks to the good initialization method of \emph{in} models which 
benefits a lot from the pretrained backbone, 
it turns out the model works not bad on these datasets. 
In Table \ref{table:kin_comp} and \ref{table:ucf101_comp}, 
our model can outperform many other 
action recognition methods. Specifically, we outperform I3D 
\cite{Carreira2017QuoVA} on Kinetics, showing our ABM-C module can work 
complementarily with the 3D architecture. 
The model fine-tuned on UCF101 and HMDB51 
are also competitive, achieving 95.1\% and 72.7\% accuracy respectively 
using only RGB inputs.

\subsection{Keyframe Selection Visualization}
\begin{figure}[t]
\begin{center}
   \includegraphics[width=\linewidth]{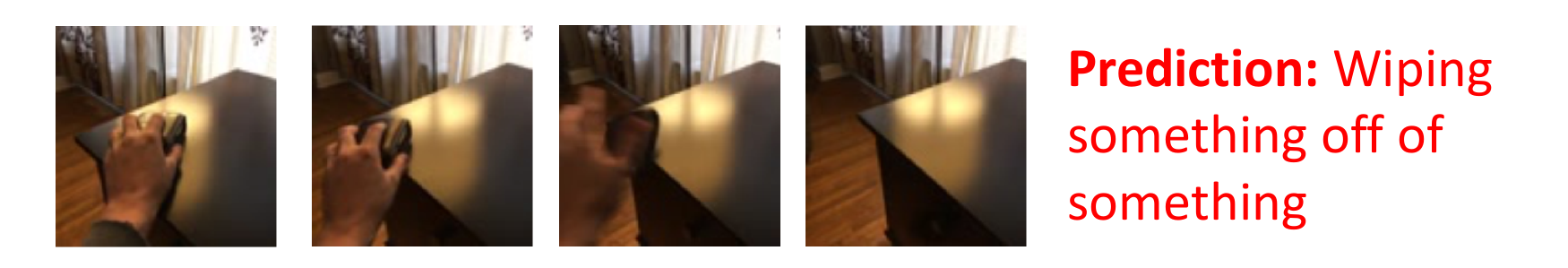}\vspace{-2.5ex}
\end{center}
%\vspace{-1pt}
    \caption{Our failure sample of 
    \emph{Putting something on the edge 
    of something so it is not supported and falls down}.}
\label{fig:failure}\vspace{-2.5ex}
\end{figure}

We conduct a keyframe selection experiment to qualitatively 
show that our ABM-C-\emph{top} model can more effectively 
capture fine-grained temporal moments than TRN \cite{Zhou2017TemporalRR}. 
We compare these two models because they are both post-CNN temporal models 
and have similar structures. Specifically, on Something-v1 
validation set we divide a video into 8 
segments and randomly sample one frame per segment to 
generate a candidate tuple. 
200 generated candidate tuples 
are fed to networks to get predictions, then we select the 
tuple with the highest top-1 prediction score as the keyframes for this video. 
In Fig. \ref{fig:keyframe} we show two videos in which two models selected 
different keyframes (only center 4 frames are shown since the rest 
selected frames are almost the same for both models). 
We can see our model performs better on capturing instant 
key moments in both videos: TRN focused on the falling 
moment of the objects, while our model captured the instant 
standing moment of edge supporting. 
A failure sample of our model is provided in Fig. \ref{fig:failure}. 
However the failure is caused by the wiping motion of a hand 
instead of capturing key moments, and the failure 
frequency is much lower.

%------------------------------------------------------------------------
\section{Conclusion}
We brought bilinear modules to temporal modeling, motivated by 
the temporal reasoning and fine-grained classification. The key points 
that made ABMs effective and efficient are 1. connection between MLPs and 
ABMs, and 2. static-plus-dynamic representation of frame features. 
By exploiting these two ideas, effective ABM temporal variants were 
proposed, which can work smoothly with existing deep CNNs. 
We showed in detail how subnets in deep CNNs can be converted to ABMs by 
adding the \emph{auxiliary-branch}, while keeping the pretrained 
parameters valid. Our modules were instantiated with 2D-ResNet-34 and I3D 
backbones. Additionally, we introduced snippet sampling and shifting 
inference to boost performance of sparse-frame models. 
It was shown that \emph{top} models are highly efficient while \emph{in} 
models are highly effective. 
Our models outperformed most state-of-the-art 
methods on Something-v1 and v2, 
and were also competitive for traditional action recognition tasks.
Though not explored in this paper, our modules should work with 
other techniques like attention mechanism 
\cite{Du2017RecurrentSA,Li2018VideoLSTMCA} and 
non-local module \cite{NonLocal2018}, which are remained for future works.
\vspace{-1.5ex}

\section*{Acknowledgement\vspace{-1.5ex}}
{\footnotesize This work is supported by Australian
Research Council under Projects FL-170100117, DP-180103424, DE-180101438, 
and in part by the National Natural Science Foundation of China under 
Grants 61772332.
}
%FL-170100117, Project
%DP-180103424, Project IH180100002, and Project DE180101438.

{\small
\bibliographystyle{ieee_fullname}
\bibliography{egbib}
}

\end{document}